%% file: main.tex
\pdfoutput=1
\documentclass[11pt]{article}

\usepackage[preprint]{acl}

\usepackage{microtype}

\usepackage{times}
\usepackage[T1]{fontenc}
\usepackage[utf8]{inputenc}

\usepackage{graphicx}
\usepackage{amsmath}
\usepackage{multirow}
\usepackage{tabularx}
\usepackage{booktabs}
\usepackage{float}
\usepackage{makecell}
\usepackage[table]{xcolor}
\usepackage{url}

\title{ExpArt-KG: Artwork Image Description Generation \\through Iterative Exploration of Knowledge Graphs}

\author{
  \textbf{Yuta Kato\textsuperscript{1}},\;\;\;
  \textbf{Shintaro Ozaki\textsuperscript{2}},\;\;\;
  \textbf{Kazuki Hayashi\textsuperscript{2}},\;\;\;
  \textbf{Yusuke Sakai\textsuperscript{2}},\;\;\;
\\
  \textbf{Hidetaka Kamigaito\textsuperscript{2}},\;\;\;
  \textbf{Katsuhiko Hayashi\textsuperscript{1,\;2}},\;\;\;
  \textbf{Taro Watanabe\textsuperscript{2}}
\\
\\
  \textsuperscript{1}The University of Tokyo,\;\;\;
  \textsuperscript{2}Nara Institute of Science and Technology (NAIST)
\\
  \texttt{\{ukato6209, katsuhiko-hayashi\}@g.ecc.u-tokyo.ac.jp}
\\
  \texttt{ozaki.shintaro.ou6@naist.ac.jp}
\\
  \texttt{\{hayashi.kazuki.hl4, sakai.yusuke.sr9, kamigaito.h, taro\}@is.naist.jp}
}

\begin{document}

\maketitle

\begin{abstract}
Large Vision-Language Models (LVLMs) achieve strong performance on image-grounded text generation and visual question answering.
However, it remains difficult for them to comprehensively and accurately describe the factual relations among the entities and concepts associated with the objects depicted in an image.
In this work, we propose a framework that efficiently exploits factual information from a knowledge graph via retrieval-augmented generation (RAG), with the goal of enabling LVLMs to generate detailed and accurate image explanations.
Specifically, our method alternates between answer generation and knowledge-graph retrieval, and controls the search using a correctness judgment, thereby acquiring the necessary and sufficient factual information efficiently.
We also construct a knowledge graph for the artwork domain (ExpArt-KG), in which the correspondence between images and entities is unambiguous.
Applying the proposed method to this knowledge graph, we show experimentally that it improves the level of detail of artwork explanations and reduces the retrieval cost of external knowledge while maintaining generation quality comparable to that of iterating a fixed number of times.
\end{abstract}

\section{Introduction}

Large Vision-Language Models (LVLMs)~\cite{liu2023visual,bai2025qwen2,abdin2024phi} achieve strong performance on image-grounded text generation and visual question answering~\cite{li2025surveystateartlarge,bai2023qwenvlversatilevisionlanguagemodel,dai2023instructblipgeneralpurposevisionlanguagemodels}, one instance of which is the ability to generate detailed explanations of an image.
In such settings, an LVLM is required not merely to recognize the people and objects contained in an image, but also to connect them with their surrounding knowledge.
However, it has been pointed out that LVLMs struggle to comprehensively and accurately explain the factual relations among the entities associated with what they recognize~\cite{hayashi2024artworkexplanationlargescalevision}.

A practical way to compensate for this limitation of LVLMs is retrieval-augmented generation (RAG)~\cite{lewis2020retrieval}, which utilizes external knowledge.
This framework generates answers on the basis of retrieved and reliable external information, thereby suppressing hallucinations and supplying the model with detailed knowledge.
In particular, for handling complex factual relations among entities comprehensively and accurately, it is effective to exploit a knowledge graph, whose systematic structure makes the factual relations among entities explicit.
While such an approach can extend the knowledge of a model without additional training, existing knowledge-graph-based methods fix the search depth or the number of iterations in advance, which creates a trade-off: a shallow search yields insufficient information, whereas a deep search increases the retrieval cost~\cite{li2025simpleeffectiverolesgraphs}.
Consequently, a general framework for efficiently retrieving useful factual relations and properly using them for explanation generation has not been sufficiently established.

In this work, we propose an iterative RAG framework that introduces a correctness judgment of the generated answer by a Large Language Model (LLM)~\cite{yang2025qwen3technicalreport,achiam2023gpt,comanici2025gemini,grattafiori2024llama} acting as a judge~\cite{zheng2023judgingllmasajudgemtbenchchatbot}, and that dynamically explores a knowledge graph and regenerates the answer on the basis of that judgment, aiming at detailed image explanation generation by LVLMs.
In addition, we construct a knowledge graph (ExpArt-KG) for extending the knowledge of LVLMs, targeting the artwork domain, in which a primary image and its entity are in one-to-one correspondence.

Evaluating the explanation generation ability of an LVLM using the proposed method together with ExpArt-KG under evaluation metrics that cover several viewpoints, we confirm that iterative answer generation explores the knowledge graph efficiently and improves the level of detail of the generated explanations.
We further show that iteration control based on the correctness judgment reduces the retrieval cost of external knowledge while maintaining generation quality comparable to that of iterating a fixed number of times.

\section{Proposed Method}

The proposed retrieval method iteratively performs a local exploration of the nodes corresponding to the entities contained in the query given to the LVLM, together with the generation and verification of an answer that uses the retrieved factual information. In this way, the method obtains useful factual information while keeping the retrieval cost of external knowledge low.
Following prior work in which combining LLM self-verification with RAG through iterative answering was shown to improve performance~\cite{hayashi2025iterkeyiterativekeywordgeneration}, answer verification by an LLM is introduced so as to achieve both efficient exploration and high-quality explanation generation.
The prompts used in the method are detailed in Appendix~\ref{app:prompt}.

The concrete procedure of the proposed method is as follows.

\paragraph{(1) Triple retrieval.}
Among the entities contained in the query, those that exist as nodes of the knowledge graph are extracted, and the triples containing them are retrieved.

\paragraph{(2) Triple selection.}
The triples obtained in Step 1 are ranked by TF-IDF~\cite{salton1988term}, and the top 10 triples are selected for each entity.
In this ranking, the constituents of a triple are regarded as terms, and the set of triples containing a particular entity as a document.
The detailed computation of TF-IDF is given in Appendix~\ref{app:tf-idf}.

\paragraph{(3) Answer generation.}
An augmented query is constructed by appending the triples selected in Step 2 to the original query, and it is given to the LVLM together with the target artwork (image) to generate an answer.

\paragraph{(4) Answer verification.}
The original query and the answer generated in Step 3 are given to an LLM, which verifies whether the answer is correct.
Using forced decoding, the LLM outputs either ``True'' or ``False''.
If ``True'' is obtained, the answer is taken as the final answer of the LVLM; if ``False'' is obtained, triple retrieval and answer generation are performed again.

\paragraph{(5) Triple re-retrieval.}
When ``False'' is obtained in Step 4, triples are retrieved again by the procedures of Steps 1 and 2, using either the original query or the answer generated immediately before, and the answer is then regenerated and verified through Steps 3 and 4.
This process is repeated until ``True'' is obtained or the maximum number of iterations is reached.

\section{Knowledge Graph Construction}

This section introduces a general procedure for constructing a knowledge graph suited to generating explanations of images.
Based on the textual information contained in an arbitrary question-answering dataset, entities and their relations are extracted by reference to an external knowledge base, and a knowledge graph is thereby constructed.
The concrete construction procedure is as follows.

\paragraph{(1) Extraction of node candidates.}
From the strings contained in the questions, reference explanations, and any available metadata (such as the titles of images) of a question-answering dataset, the expressions that are candidate nodes of the knowledge graph are extracted.

\paragraph{(2) Node selection.}
Among the node candidates extracted in Step 1, those that exist as titles of English Wikipedia articles are adopted as entities.
Furthermore, highly ambiguous words, for which multiple corresponding Wikidata~\cite{42240} identifiers exist, and expressions denoting general concepts are excluded from the nodes, so that the resulting knowledge graph is dense and specialized to the target domain.
This leaves only entities that are specific to, and unambiguous within, the target domain.

\paragraph{(3) Edge construction.}
For each node selected in Step 2, the corresponding Wikidata entity is referred to, and the predicates defined among them are obtained.
Defining these predicates as edges that express semantic relations between nodes yields a knowledge graph in which the factual relations among entities are structured.

In this work, we applied this procedure to ExpArt~\cite{hayashi2024artworkexplanationlargescalevision} to construct ExpArt-KG, a knowledge graph about artworks.

\input{score_table}

\section{Experimental Setup}

\paragraph{Models.}
We used Qwen3-VL~\cite{bai2025qwen3vltechnicalreport} as the LVLM that generates image explanations, and Qwen3~\cite{yang2025qwen3technicalreport} as the LLM that verifies answers (we refer to it as the validator LLM).

\paragraph{Dataset.}
From the test data of ExpArt,\footnote{\url{https://huggingface.co/datasets/naist-nlp/ExpArt}} we targeted the instances whose image URLs\footnote{\url{https://commons.wikimedia.org/wiki/Main_Page}} were valid as of December 2025 and whose answers contained at least one entity of the constructed ExpArt-KG. From these we sampled approximately 25\%, which we used for evaluation.
Details are given in Appendix~\ref{app:dataset}.

\paragraph{Input settings.}
In this experiment, we used the following two question settings for each image.

\noindent (1) \textbf{With Title}:
The LVLM generates an explanation from the image and a question that contains the title.
The validator LLM judges correctness on the basis of the question containing the title and the answer.

\noindent (2) \textbf{Without Title}:
When generating the image explanation, the LVLM is given the image and a question that does not contain the title.
The validator LLM is likewise given only the question without the title and the answer, so that the correctness judgment is made solely on the basis of the logical consistency of the answer.

Through these settings, we examine the difference in generation and verification performance depending on the presence of the title, as well as the effect of the image title on the correctness judgment of an answer.

\paragraph{Knowledge graph retrieval.}
We define three variants of TF-IDF-based triple ranking:
(1) treating predicates as terms and using their scores (PID),
(2) treating adjacent entities as terms and using their scores (QID), and
(3) treating both as terms and using the sum of the two scores (PID-QID).
Details are given in Appendix~\ref{app:tf-idf}.

\paragraph{RAG settings.}
To examine the effectiveness of ExpArt-KG and of the proposed method (RAG-Validate), we compared RAG-Validate with a method that generates an answer only once without RAG (Baseline) and a method that uses RAG and repeats answer generation five times (RAG-Loop5).

\paragraph{Evaluation metrics.}
For evaluation, we used BLEU~\cite{papineni-etal-2002-bleu}, ROUGE~\cite{lin-2004-rouge}, and BERTScore~\cite{zhang2020bertscoreevaluatingtextgeneration}, which are standard in natural language generation tasks, in order to measure the lexical and semantic similarity between the generated text and the reference explanation.
In addition, to measure the level of detail of explanations about artworks, we used Entity Coverage, Entity F1, and Entity Cooccurrence~\cite{hayashi2024artworkexplanationlargescalevision}.
The corresponding formulas are given in Appendix~\ref{app:metrics}.

\noindent (1) \textbf{Entity Coverage}
evaluates the extent to which the generated text covers the artwork-related entities in the reference explanation, under two settings: exact match and partial match.

\noindent (2) \textbf{Entity F1}
compares the artwork-related entities contained in the generated text and in the reference explanation, and evaluates the frequency of entity occurrences and their appropriate usage.

\noindent (3) \textbf{Entity Cooccurrence}
focuses on the cooccurrence relations among entities across the whole text and evaluates the coverage of cooccurrences.
It adopts a length penalty analogous to the brevity penalty of BLEU~\cite{papineni-etal-2002-bleu} to prevent overrating redundant explanations.

\begin{figure}[!t]
    \centering
    \includegraphics[width=\linewidth]{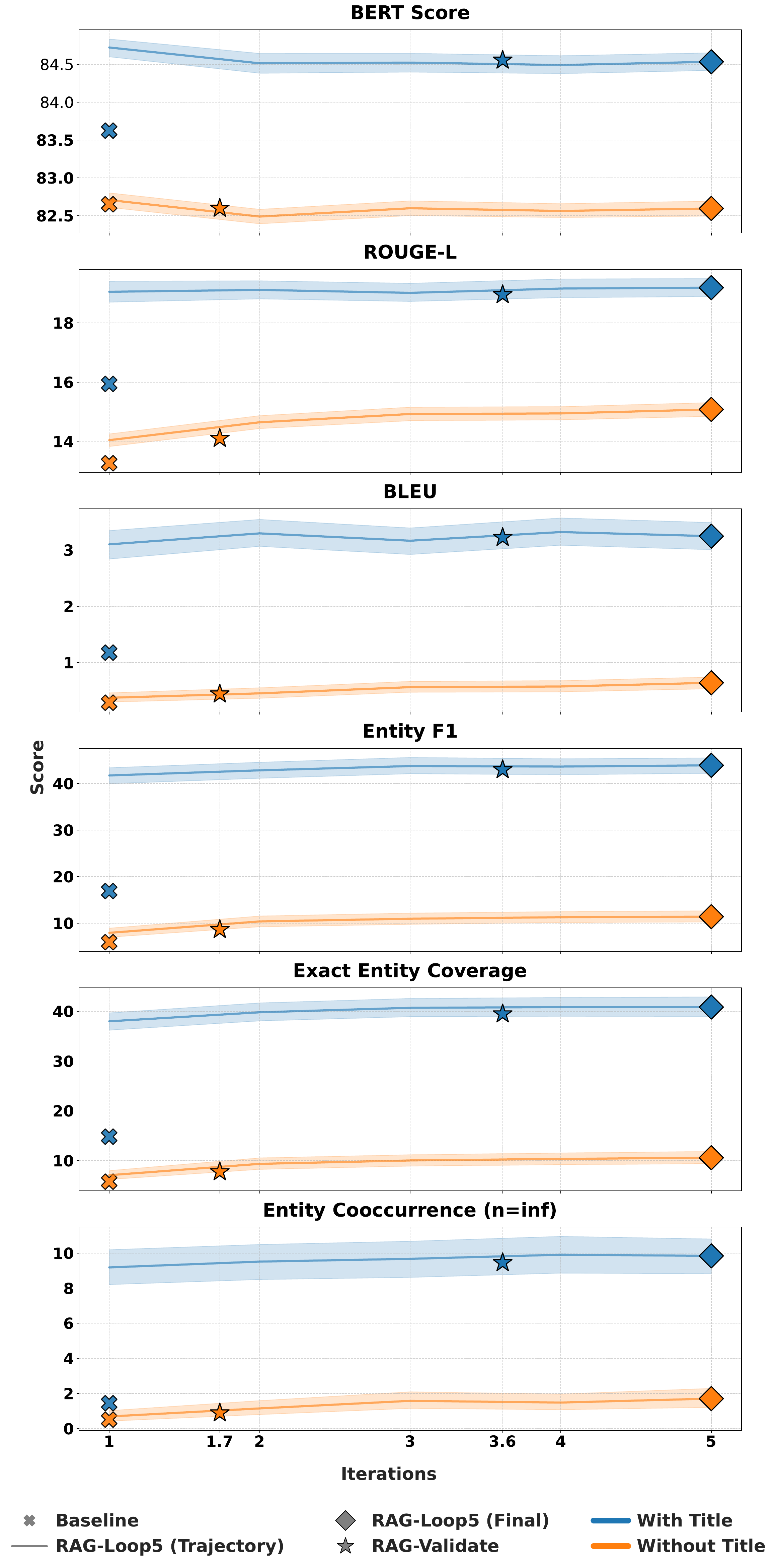}
    \caption{Scores obtained at each iteration of RAG-Loop5, together with those of the Baseline and RAG-Validate, under the QID setting. The horizontal axis denotes the number of iterations and the vertical axis denotes the score. Solid lines show the transition of the RAG-Loop5 scores, and the shaded regions indicate their 95\% confidence intervals. Crosses mark the Baseline, diamonds the final RAG-Loop5 scores, and stars RAG-Validate, plotted at its average number of iterations (3.6 with the title and 1.7 without it).}
    \label{fig:iteration_graph}
\end{figure}

\section{Results and Analysis}

Table~\ref{tab:score_table} reports the scores of the evaluation metrics under each setting.
For all metrics except BERTScore, RAG-Validate and RAG-Loop5 improved over the Baseline.
While BERTScore stays at a high level regardless of whether RAG is used, marked improvements are observed for Entity F1, Entity Coverage, and Entity Cooccurrence.
These results indicate that, although the Baseline already produces grammatically fluent text, using ExpArt-KG and the proposed method substantially improves the accuracy and comprehensiveness of the entities contained in the explanations, and can raise the level of detail of the explanations with respect to the depicted subject and its surrounding knowledge.
Regarding the way TF-IDF scores are computed, the largest improvement in the With Title setting was observed when only adjacent entities were used as the constituent terms.
This suggests that using the entities of a triple, rather than its predicates, for retrieval improves the ability to identify highly specific entities.
Moreover, for questions containing the title of the artwork, RAG-Validate attains scores comparable to RAG-Loop5.

Figure~\ref{fig:iteration_graph} shows the transition of scores against the number of iterative answer generations for the Baseline, RAG-Loop5, and RAG-Validate.
In the Without Title setting, the score of the first answer is markedly lower than in the With Title setting, but the score was confirmed to increase continuously as RAG-Loop5 repeated answer generation.
In the With Title setting, on the other hand, the score also increases, but the magnitude of the increase tends to diminish after a certain number of iterations.
This suggests that traversing the nodes of the knowledge graph multiple times makes it possible to gradually acquire factual information useful for answer generation.
In particular, the fact that score improvements were confirmed even when the initial answer contained few related entities and thus offered few cues for exploration supports the view that iterative retrieval functions as multi-hop reasoning over the knowledge graph and effectively explores entities that are useful for answer generation.

Furthermore, the average number of iterative answer generations of RAG-Validate is 3.6 in the With Title setting; despite requiring fewer iterations than the fixed-count RAG-Loop5, it maintains answer scores at a level comparable to RAG-Loop5.
This result shows that the proposed method---which judges the correctness of the answer generated by the LVLM with a validator LLM and terminates the iteration once the answer is deemed valid---avoids the unnecessary exploration that arises when a fixed number of iterations is always performed, and thus reduces the retrieval cost of external knowledge without degrading generation quality.

Finally, whereas RAG-Validate maintains scores comparable to RAG-Loop5 in the With Title setting, it obtains lower scores than RAG-Loop5 in the Without Title setting.
The likely cause of this degradation is that the validator lacks the knowledge that serves as the basis for the correctness judgment, so that it fails to properly reject incorrect answers and terminates the exploration prematurely.
This indicates that verification in RAG-Validate is based not on the mere fluency of the generated text, but on the factual consistency between the input title and the content of the answer.

\section{Conclusion}

Aiming to improve the level of detail of image explanations, we proposed a retrieval-augmented generation method that makes effective use of a knowledge graph.
The proposed method iteratively performs correctness judgment and regeneration of the generated answer and dynamically explores the knowledge graph, which makes it possible to efficiently acquire entities that are useful for explanation generation.
In addition, we constructed ExpArt-KG, a knowledge graph consisting of entities related to artworks.
Our experimental results confirm that, by dynamically controlling the number of iterations according to the validity of the answer, the proposed method reduces the total number of iterations while maintaining generation quality comparable to that of iterating a fixed number of times, thereby achieving both high-quality explanation generation and a reduced retrieval cost for external knowledge.

\section*{Acknowledgments}
This work was supported by JSPS KAKENHI Grant Numbers JP23K28148 and JP24K02993.

\bibliography{custom}

\appendix

\section{Appendix}
\label{sec:appendix}

\setlength{\abovedisplayskip}{3pt}
\setlength{\belowdisplayskip}{3pt}
\setlength{\abovedisplayshortskip}{2pt}
\setlength{\belowdisplayshortskip}{2pt}

\subsection{Prompts}
\label{app:prompt}
The prompts used in the proposed method are shown in Table~\ref{tab:prompts}.

\input{prompt_table}

\subsection{Details of the Models Used}
\label{app:models}
Table~\ref{tab:model_ids} shows the correspondence between the abbreviations used in this paper and the Hugging Face model IDs.

\begin{table}[t]
    \centering
    \renewcommand{\arraystretch}{0.8}
    \small
    \begin{tabularx}{\linewidth}{l X}
        \toprule
        \textbf{Abbreviation} & \textbf{Hugging Face Model ID} \\
        \midrule
        Qwen3-VL & Qwen/Qwen3-VL-8B-Instruct \\
        Qwen3 & Qwen/Qwen3-4B-Instruct-2507 \\
        \bottomrule
    \end{tabularx}
    \caption{Correspondence between the models used and their Hugging Face IDs.}
    \label{tab:model_ids}
\end{table}

\subsection{Dataset Statistics}
\label{app:dataset}
Table~\ref{tab:data_stats} shows the number of questions at each processing stage of the dataset.
Note that, by the specification of the data, each question comes in two settings depending on whether the title is included.

\begin{table}[t]
    \centering
    \renewcommand{\arraystretch}{0.8}
    \small
    \begin{tabular}{lr}
        \toprule
        \textbf{Breakdown} & \textbf{\# Questions} \\
        \midrule
        ExpArt test data & 5,227 \\
        After filtering & 4,823 \\
        After sampling (evaluation data) & 1,199 \\
        \bottomrule
    \end{tabular}
    \caption{Dataset statistics.}
    \label{tab:data_stats}
\end{table}

\subsection{Details of TF-IDF-Based Triple Selection}
\label{app:tf-idf}

In our method, the set of triples connected to an entity $e$ is regarded as a document $D_e$, and a constituent of a triple $T \in D_e$ (its predicate $p$ or its adjacent node $q$) is regarded as a term $t$.
Let $\mathcal{E}$ be the set of all entities of the knowledge graph and $N = |\mathcal{E}|$ its size, let $\mathrm{Terms}(D)$ denote the set of terms occurring in the triples of a document $D$, let $f_{t,e}$ be the number of times the term $t$ occurs in $D_e$, and let $n_t = |\{e' \in \mathcal{E} : t \in \mathrm{Terms}(D_{e'})\}|$ be the number of entities whose document contains $t$.
The TF-IDF value $w(t, D_e)$, which represents the importance of the term $t$, is then defined as follows, where logarithmic normalization and smoothing are applied for numerical stability.
\begin{equation}
    \small
    w(t, D_e) = \log(1 + f_{t,e}) \cdot \log \left( \frac{N + 1}{n_t + 1} \right)
\end{equation}
The score $S(T)$ of a triple $T$ whose predicate is $p$ and whose adjacent node is $q$ is computed as follows under each setting.
(1) \textbf{PID}: $S(T) = w(p, D_e)$,
(2) \textbf{QID}: $S(T) = w(q, D_e)$,
(3) \textbf{PID-QID}: $S(T) = w(p, D_e) + w(q, D_e)$.
Based on this score, we select the top-ranked triples for each entity.

\subsection{Definitions of the Entity Evaluation Metrics}
\label{app:metrics}

Let a generated explanation consisting of $k$ sentences be $G = \{g_1, \dots, g_k\}$ and a reference explanation consisting of $m$ sentences be $R = \{r_1, \dots, r_m\}$.
Let the sets of entities extracted from these texts be $E_G = \mathrm{Entity}(G)$ and $E_R = \mathrm{Entity}(R)$, and let $|G|$ and $|R|$ denote their respective total numbers of tokens.
We define the length penalty $\mathrm{LP}(G, R)$ for redundant generated text as follows.
\begin{equation}
    \small
    \mathrm{LP}(G, R) = \exp\left(-\max\left(0.0, \ \frac{|G|}{|R|} - 1\right)\right)
\end{equation}

\noindent Each evaluation metric is computed as follows.
In what follows, $\mathrm{Cov}(X, Y)$ denotes the proportion of the elements of $Y$ that are covered by $X$.

\noindent (1) \textbf{Entity Coverage (EC)}:
the proportion of the entities contained in $R$ that are covered by $G$, computed by the following equation.
The longest common subsequence is used to compute the partial-match coverage rate.
\begin{equation}
    \small
    \mathrm{EC}(G, R) = \mathrm{Cov}(E_G, E_R)
\end{equation}

\noindent (2) \textbf{Entity F1 (EF1)}:
the harmonic mean of the precision $\mathit{Prec}$ and the recall $\mathit{Rec}$ based on the frequency of entity occurrences.
Letting $\#(e, \cdot)$ be the number of occurrences of entity $e$ in the target text and $\mathrm{Count}_{\text{clip}}(e, G, R) = \min(\#(e, G), \#(e, R))$, they are defined by the following equations.
\begin{gather}
    \small
    \mathrm{EF}_1 = \frac{2 \times \mathit{Prec} \times \mathit{Rec}}{\mathit{Prec} + \mathit{Rec}} \\
    \small
    \mathit{Prec} = \frac{\sum_{e \in E_G} \mathrm{Count}_{\text{clip}}(e, G, R)}{\sum_{e \in E_G} \#(e, G)} \\
    \small
    \mathit{Rec} = \frac{\sum_{e \in E_R} \mathrm{Count}_{\text{clip}}(e, G, R)}{\sum_{e \in E_R} \#(e, R)}
\end{gather}

\noindent (3) \textbf{Entity Cooccurrence (ECooc)}:
letting $\mathrm{Co}_n(\cdot)$ be a function that returns the pairs of entities cooccurring within a context window consisting of a sentence and its surrounding $n$ sentences, it is computed by the following equation.
We used the sentence splitter of NLTK to split the text into sentences.
\begin{equation}
    \footnotesize
    \mathrm{ECooc}(G, R) = \mathrm{LP}(G, R) \times \mathrm{Cov}(\mathrm{Co}_n(G), \mathrm{Co}_n(R))
\end{equation}

\end{document}

%% file: score_table.tex
\begin{table*}[t]
\centering
\resizebox{\textwidth}{!}{%
\setlength{\tabcolsep}{5pt}
\begin{tabular}{l l c ccc c c cc cccc c}
\toprule
\multirow{2}{*}{\textbf{RAG setting}} & \multirow{2}{*}{\textbf{TF-IDF}} & \multirow{2}{*}{\textbf{BLEU}} & \multicolumn{3}{c}{\textbf{ROUGE}} & \multirow{2.5}{*}{\shortstack{\textbf{BERT} \\ \textbf{Score}}} & \multirow{2.5}{*}{\shortstack{\textbf{Entity} \\ \textbf{F1 ($\uparrow$)}}} & \multicolumn{2}{c}{\textbf{Entity Coverage ($\uparrow$)}} & \multicolumn{4}{c}{\textbf{Entity Cooccurrence ($\uparrow$)}} & \multirow{2.5}{*}{\shortstack{\textbf{Output} \\ \textbf{length}}} \\
\cmidrule(lr){4-6} \cmidrule(lr){9-10} \cmidrule(lr){11-14}
 & &  & \textbf{1} & \textbf{2} & \textbf{L} &  &  & \textbf{Exact} & \textbf{Partial} & \textbf{n=0} & \textbf{n=1} & \textbf{n=2} & \textbf{n=$\infty$} &  \\
\midrule
\multicolumn{15}{c}{\textbf{With Title}} \\
RAG-Loop5 & PID & 3.22 & 30.73 & 9.02 & 19.02 & 84.51 & 41.47 & 38.52 & 45.07 & 7.93 & 8.77 & 8.89 & 8.90 & 121.6 \\
RAG-Loop5 & PID-QID & 3.12 & 30.89 & 9.01 & 18.97 & 84.53 & 43.24 & 39.95 & 46.36 & 8.16 & 9.11 & 9.53 & 9.45 & 121.2 \\
RAG-Loop5 & QID & 3.22 & \textbf{31.17} & \textbf{9.11} & \textbf{19.20} & 84.54 & \textbf{43.85} & \textbf{40.73} & \textbf{47.10} & \textbf{8.60} & \textbf{9.61} & \textbf{9.92} & \textbf{9.79} & 123.8 \\
\cmidrule(lr){2-15}
\rowcolor{cyan!5}
RAG-Validate & PID & 3.15 & 30.55 & 8.92 & 18.90 & 84.53 & 41.89 & 37.93 & 44.45 & 7.89 & 8.43 & 8.46 & 8.41 & 116.3 \\
\rowcolor{cyan!5}
RAG-Validate & PID-QID & 3.18 & 30.66 & 8.96 & 18.92 & 84.55 & 42.17 & 38.69 & 45.34 & 8.47 & 8.71 & 9.01 & 8.95 & 118.1 \\
\rowcolor{cyan!5}
RAG-Validate & QID & \textbf{3.22} & 30.85 & 9.07 & 18.96 & \textbf{84.56} & 42.96 & 39.46 & 45.99 & 8.53 & 9.19 & 9.52 & 9.45 & 120.1 \\
\cmidrule(lr){2-15}
\rowcolor{gray!10} Baseline & -- & 1.18 & 24.89 & 5.53 & 15.94 & 83.62 & 16.93 & 14.82 & 22.67 & 1.76 & 1.55 & 1.53 & 1.44 & 91.3 \\
\midrule
\multicolumn{15}{c}{\textbf{Without Title}} \\
RAG-Loop5 & PID & 0.54 & 23.92 & 3.95 & 14.88 & 82.61 & 11.29 & 10.02 & 17.33 & 1.05 & 1.20 & 1.27 & 1.24 & 119.6 \\
RAG-Loop5 & PID-QID & 0.60 & 24.02 & \textbf{4.06} & 14.91 & 82.64 & \textbf{11.75} & \textbf{10.66} & \textbf{17.80} & \textbf{1.49} & \textbf{1.68} & \textbf{1.78} & \textbf{1.73} & 120.7 \\
RAG-Loop5 & QID & \textbf{0.61} & \textbf{24.18} & 3.98 & \textbf{14.94} & 82.60 & 11.12 & 10.33 & 17.62 & 1.41 & 1.60 & 1.73 & 1.66 & 123.8 \\
\cmidrule(lr){2-15}
\rowcolor{cyan!5}
RAG-Validate & PID & 0.43 & 22.58 & 3.51 & 14.14 & 82.62 & 8.59 & 7.94 & 15.22 & 0.76 & 0.95 & 0.97 & 0.95 & 102.7 \\
\rowcolor{cyan!5}
RAG-Validate & PID-QID & 0.43 & 22.51 & 3.52 & 14.08 & 82.61 & 8.63 & 7.76 & 15.21 & 0.77 & 0.75 & 0.78 & 0.78 & 103.8 \\
\rowcolor{cyan!5}
RAG-Validate & QID & 0.44 & 22.52 & 3.52 & 14.10 & 82.60 & 8.65 & 7.77 & 15.08 & 0.69 & 0.86 & 0.90 & 0.89 & 104.9 \\
\cmidrule(lr){2-15}
\rowcolor{gray!10} Baseline & -- & 0.29 & 20.59 & 3.18 & 13.26 & \textbf{82.65} & 5.95 & 5.79 & 12.55 & 0.54 & 0.52 & 0.50 & 0.51 & 82.1 \\
\bottomrule
\end{tabular}
}
\caption{Scores obtained under the various settings of our experiments. \textbf{Bold} indicates the best value for each evaluation metric. Among the shaded rows, cyan denotes the proposed method and gray denotes the Baseline.}
\label{tab:score_table}
\end{table*}

%% file: prompt_table.tex
\begin{table}[t]
    \centering
    \small
    \renewcommand{\arraystretch}{0.8}
    \begin{tabularx}{\linewidth}{l X}
        \toprule
        \textbf{Usage} & \textbf{Prompt} \\
        \midrule
        \makecell[lt]{\textbf{LVLM} \\ \textbf{w/o KG}} & \textbf{System:} You are an assistant helping with visual question answering. \newline \textbf{User:} Question: \textcolor{red}{\{question\}} \newline Answer the question based on the provided image. \newline Do not include any additional text other than the answer itself. \\
        \midrule
        \makecell[lt]{\textbf{LVLM} \\ \textbf{w/ KG}} & \textbf{System:} You are an assistant helping with visual question answering using external knowledge. \newline \textbf{User:} Below are factual triples retrieved from a knowledge graph to help answer the question. \newline Each triple is formatted as: [SUBJECT, PREDICATE, OBJECT] \newline Retrieved Facts: \textcolor{red}{\{triplets\}} \newline Question: \textcolor{red}{\{question\}} \newline Answer the question by integrating the image, your internal knowledge, and any retrieved facts that are relevant to the question. \newline Do not include any additional text other than the answer itself. \\
        \midrule
        \makecell[lt]{\textbf{Validator} \\ \textbf{LLM}} & \textbf{System:} You are an assistant that validates answers. \newline \textbf{User:} Question: \textcolor{red}{\{question\}} \newline Answer: \textcolor{red}{\{answer\}} \newline Judge whether the provided answer is correct and addresses the question. \newline Respond with only "True" or "False". Do not provide any additional explanation or text. \\
        \bottomrule
    \end{tabularx}
    \caption{Prompts used in the proposed method.}
    \label{tab:prompts}
\end{table}